\documentclass{article}

\usepackage{arxiv}

\usepackage[utf8]{inputenc} 
\usepackage[T1]{fontenc}    
\usepackage{hyperref}       
\usepackage{url}            
\usepackage{booktabs}       
\usepackage{amsfonts}       
\usepackage{nicefrac}       
\usepackage{microtype}      
\usepackage{lipsum}
\usepackage{colortbl,float}
\usepackage{caption}
\usepackage{amsmath,graphicx}
\usepackage{listings}
\usepackage{subcaption}
\usepackage{enumitem}
\usepackage[numbers]{natbib}
\graphicspath{ {imgs/} }

\usepackage[ruled]{algorithm2e} 

\newcommand{\nint}[1]{\ensuremath{\lfloor#1\rceil}}

\title{\huge An Optimized Recurrent Unit for Ultra-Low-Power Keyword Spotting}

\author{
  Justice Amoh\\
  Thayer School of Engineering\\
  Dartmouth College\\
  Hanover, NH 03755 \\
  \texttt{justice.amoh.jr.th@dartmouth.edu} \\
   \And
  Kofi M. Odame\\
  Thayer School of Engineering\\
  Dartmouth College\\
  Hanover, NH 03755 \\
  \texttt{kofi.m.odame@dartmouth.edu} \\
}

\begin{document}
\maketitle

\begin{abstract}
There is growing interest in being able to run neural networks on sensors, wearables and internet-of-things (IoT) devices. However, the computational demands of neural networks make them difficult to deploy on resource-constrained edge devices.

To meet this need, our work introduces a new recurrent unit architecture that is specifically adapted for on-device low power acoustic event detection (AED). The proposed architecture is based on the gated recurrent unit (`GRU' -- introduced by \citet{Cho2014}) but features optimizations that make it implementable on ultra-low power micro-controllers such as the Arm Cortex M0+. 

Our new architecture, the Embedded Gated Recurrent Unit (eGRU) is demonstrated to be highly efficient and suitable for short-duration AED and keyword spotting tasks. A single eGRU cell is 60$\times$ faster and 10$\times$ smaller than a GRU cell. Despite its optimizations, eGRU compares well with GRU across tasks of varying complexities. 

The practicality of eGRU is investigated in a wearable acoustic event detection application. An eGRU model is implemented and tested on the Arm Cortex M0-based Atmel ATSAMD21E18 processor. The Arm M0+ implementation of the eGRU model compares favorably with a full precision GRU that is running on a workstation. The embedded eGRU model achieves a classification accuracy 95.3\%, which is only 2\% less than the full precision GRU.

\end{abstract}

\section{Introduction}
\label{sec:intro}
Deep neural networks are a powerful way to extract information from noisy, raw sensor data that is collected in an unconstrained real-world environment. This approach has been used successfully in computer vision \cite{howard2017mobilenets}, computer audition \cite{hannun2014deep} and physiological monitoring \cite{ravi2017deep}. Since deep neural networks are generally memory- and computationally-intensive, they are typically implemented on high-end cloud compute servers. However, transmitting raw data from the sensor to the cloud has negative implications for battery life \cite{miettinen2010energy}, real-time responsiveness \cite{dillon2010cloud} and data privacy \cite{takabi2010security}. It also demands a reliable communications connection to the cloud which is not always possible. These challenges can be avoided by implementing the deep neural networks directly onto sensors.

\begin{figure}[t]
    \begin{subfigure}[b]{0.32\textwidth}
        \includegraphics[width=\textwidth]{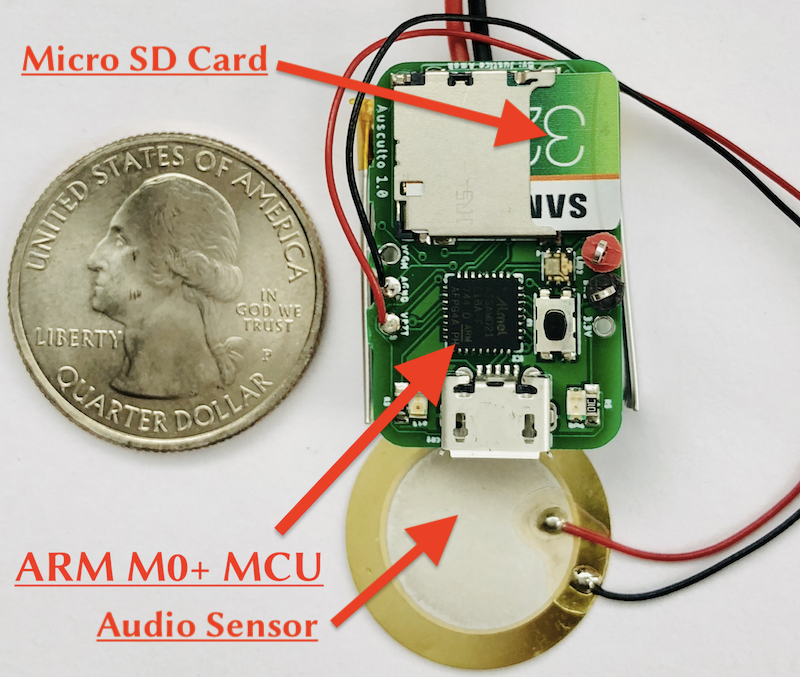}
        \caption{Our wearable cough detection device.}
        \label{fig:ausculto}
    \end{subfigure}\hfill
    \begin{subfigure}[b]{0.65\textwidth}
        {\renewcommand{\arraystretch}{1.5}
        \begin{tabular}{|l|r|r|r|r|}
        \hline
        \multicolumn{1}{|c|}{\textbf{Device}} & \multicolumn{1}{c|}{\textbf{Size}} & \multicolumn{1}{c|}{\textbf{Power}} & \multicolumn{1}{c|}{\textbf{Computations}} & \multicolumn{1}{c|}{\textbf{RAM}} \\ \hline
        Drive Px2     & $\sim$9.6$\times$9.2"  & 30,000 mW      & 20 TFLOPs             & 16 GB        \\
        Jetson Tx2    & $\sim$3.5$\times$2.5"  & 7,500 mW       & 2 TFLOPs              & 8 GB         \\
        iPhone X      & 5.6$\times$2.8"  & 1,000 mW       & 1 GFLOPs              & 3 GB         \\
        Freedom K64F  & 3.2$\times$2.1"  & 500 mW         & 1 MFLOPs              & 256 kB       \\
        \textbf{Ausculto (ours)}   & \textbf{1.1$\times$0.7"}  & \textbf{40 mW}          & \textbf{0.1 MFLOPs}            & \textbf{32 kB}        \\ \hline
        \end{tabular}}
        \vspace{10pt}
        \caption{Table of computational resources available on various portable systems.}
        \label{fig:resources}
    \end{subfigure}
    \caption{\ref{fig:ausculto} Our wearable device for detecting coughs and adventitious sounds is about the size of a thumb drive. It features the Atmel ATSAMD21E18 \cite{ATSAMD21}, an ultra-low power Arm Cortex M0+ micro-controller. Embedding a neural network to this device is difficult because of its size and power constraints. \ref{fig:resources} Compared to our device, portable systems that run neural networks like Nvidia's Drive and the iPhone X are larger and have enormous power requirements. Even devices like Freedom K64F recommended for uTensor light-weight library still need $10\times$ more power than available on our ultra-low power system.}
    \label{fig:devices}
\end{figure}

Two approaches to realizing neural networks on sensors, or “edge” devices \cite{teerapittayanon2017distributed} are (1) customized hardware processors like the Nvidia Drive Px2 and the A11 Bionic Chip \cite{nvidia_drive,apple_2017}, and (2) light-weight software libraries like uTensor and CMSIS-NN \cite{Tan2017,Lai2018}. Unfortunately, these approaches are inadequate for wearable applications like the recently-introduced cough-detection device (see Fig. \ref{fig:devices})  \cite{amoh2016deep,amoh2015deepcough,amoh2013technologies,amohmobile}. To promote long-term use, wearable devices have stringent size requirements that translate to a small battery (often the single largest component) and a correspondingly small power budget, on the order of 10 mW. Custom neural network processors like the Jetson Tx2 would drain such a battery in less than a minute. An ultra-low-power micro controller unit (MCU) \cite{ti:msp430,stm:STM32L010C6,siliconlabs:C8051F96x,microchip:PIC16LF1509} could stretch the limited power budget further, but it only provides a limited amount of memory and computational resources; even light-weight libraries have some minimum hardware requirements that cannot be met by the sparse resources of an ultra-low-power MCU.

To address these challenges, we developed a novel recurrent neural network that is specifically adapted for implementation on the Arm M0+ \cite{Armltd} class of ultra-low-power MCUs. In particular, we introduce a new architecture for the recurrent unit which is optimized for keyword spotting tasks like wearable cough detection. This paper presents details on the architecture, training scheme and hardware implementation of the proposed recurrent cell. We also present experimental results that show our architecture requires $12\times$ less memory and $60\times$ less computational time than comparable conventional recurrent units.

\section{Related Work}
\label{sec:related}
Neural network optimizations similar to those proposed in this work have previously been considered in isolation by other researchers. For instance, recent efforts to optimize the GRU architecture include the removal of one of its gates. \citet{Zhou2016} achieved this single gate architecture by coupling the update and reset gates into a single forget gate. The resulting `minimal' gated unit (MGU) directly effects both cell update and reset using the same gate. \citet{Ravanelli2017} discarded the reset gate altogether, highlighting that it is potentially redundant in speech recognition where input signals evolve slowly. Our work investigates this further in AED tasks where events are sudden and infrequent, in sharp contrast to speech recognition. Furthermore, our proposed cell architecture replaces the sigmoid and hyperbolic tangent activation functions with more efficient softsign variants.

Another common area of optimization is weight quantization. In previous works \cite{Han2015,Wu2016}, 8-bit weight quantization was shown to drastically reduce network size without significant loss in accuracy. However, these approaches focused solely on convolutional and fully-connected networks. They also required the storage of a codebook for decoding weights at run-time. With regards to recurrent networks, \cite{Ott2016} demonstrated that 2-bit weight quantization is possible in GRU, with only a slight loss in accuracy. However, the combined effect of such a low precision quantization and other optimizations like a single gate architecture remained unknown.  

Using solely integer arithmetic in neural networks has also been studied in prior works \cite{Han2015,Gupta2015,Chen2015}. \citet{Courbariaux2014} found that in a fully connected network, a dynamic point numeric format yields much better results than 20-bit fixed point. Another work demonstrates that 32-bit fixed point formats are effective in convolutional networks \cite{Chen2015}. Once again, all these efforts consider fully-connected and convolutional architectures. Since recurrent architectures are fundamentally different and much more difficult to train than other architectures \cite{pascanu2013difficulty}, it is worthwhile to investigate fixed point arithmetic in recurrent neural networks.

To date, no one has ever simultaneously applied all of these modifications to a GRU and validated that they yield a usable network that is implementable on a low power, low resource MCU like the Arm Cortex M0+. This work bridges that gap.

\section{Embedded Gated Recurrent Unit}
\label{sec:eGRU}
Previous work showed that a network of gated recurrent units (GRU) \cite{Cho2014} outperforms classical approaches like Hidden Markov Models on an acoustic event detection task when evaluated in a noisy, non-ideal environment \cite{amoh2016deep}. But the GRU network described in \cite{amoh2016deep} is too memory- and computationally-expensive to be implemented on a low-resource MCU like the Arm M0+.

To meet the resource constraints of a low power MCU like the M0+, we propose a new recurrent architecture: the embedded Gated Recurrent Unit (eGRU). It is based on the traditional GRU, but with four major modifications: (1) a single gate mechanism, (2) faster activation functions, (3) 3-bit exponential weight quantization and (4) fixed point arithmetic in all network operations. These modifications lead to massive reductions in memory and computations in the recurrent cell, making it feasible to run eGRU networks on our target device. Below, we discuss the features of the eGRU in detail.

\subsection{Single Gate Mechanism}
A practical idea for optimizing recurrent units is the removal of gates. Modern recurrent cell architectures like the Long Short-Term Memory (LSTM) unit \cite{Hochreiter1997} and GRU are characterized by gating mechanisms that regulate information flow in and out of the cell's memory. For instance, the LSTM cell has 3 gates: two for controlling the cell's input and output, and a third for forgetting or resetting the cell's internal state. Since gates are implemented using weights and activations, omitting a gate reduces the required memory and computations in a recurrent cell. Accordingly, GRU was introduced as an optimized form of LSTM by reducing the number of gates from 3 to 2: the update and reset gates. 

In the same vein, GRU can also be optimized further by discarding yet another gate. \citet{Zhou2016} accomplished this very `minimal' gated unit by using a single gate for both resetting and updating the cell's internal state. \citet{Ravanelli2017} extended that work further by highlighting a redundancy between the two gates. They deduced that in applications like speech recognition where signals change slowly, reset gates are unnecessary and can be omitted altogether. However, in applications where events of interest are abrupt and isolated (eg. detecting cough sounds), the assumption by \citet{Ravanelli2017} that state resets are irrelevant does not hold. In fact, we found that without state resets, recurrent units in our application are unable to recover from large impulse signals. Thus, for keyword spotting applications, we can eliminate the reset gate and rely on only the update gate.

\begin{figure}[t]
    \begin{subfigure}[b]{0.48\textwidth}
        \includegraphics[width=\textwidth]{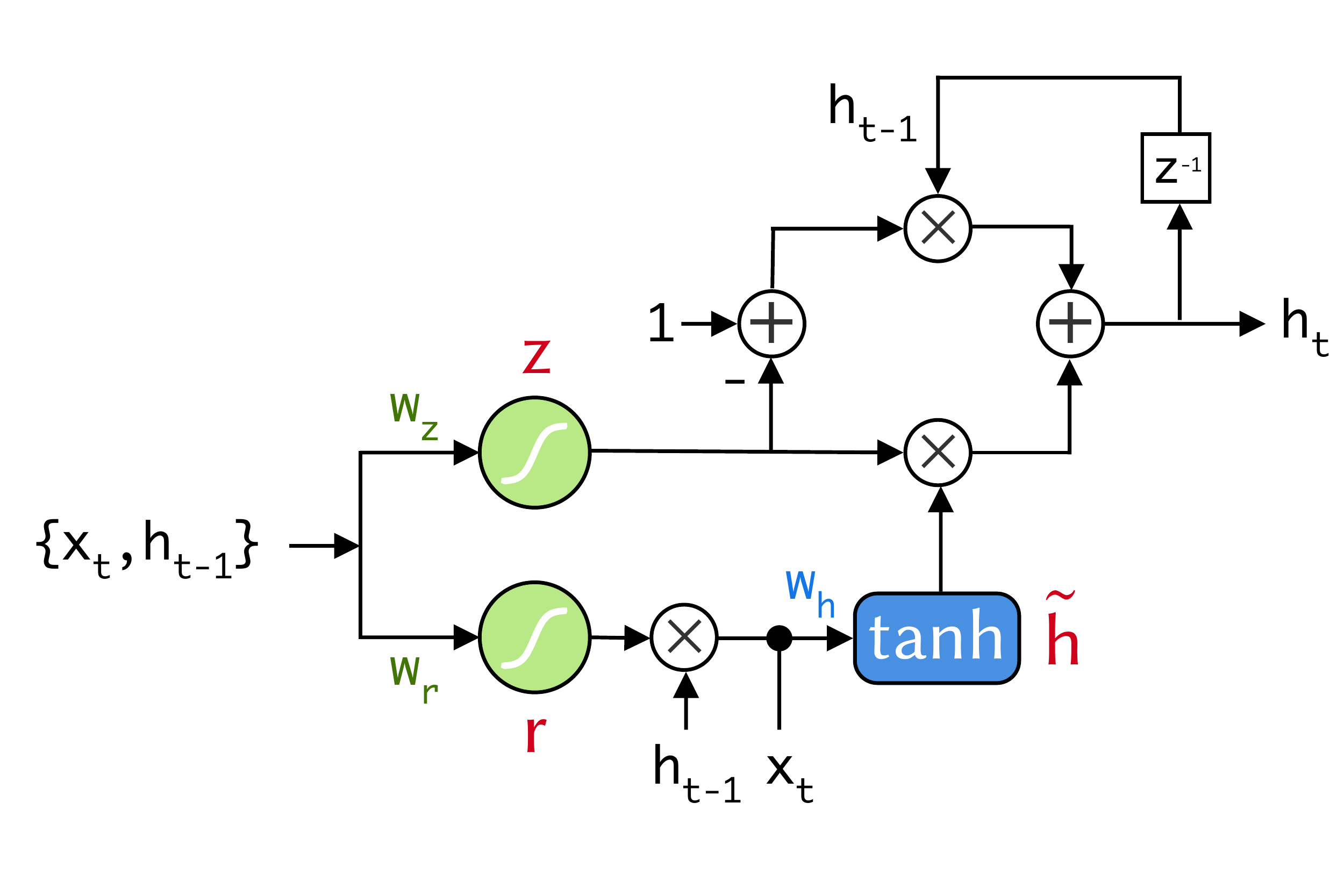}
        \caption{GRU cell.}
        \label{fig:grucell}
    \end{subfigure}\hfill
    \begin{subfigure}[b]{0.48\textwidth}
        \includegraphics[width=\textwidth]{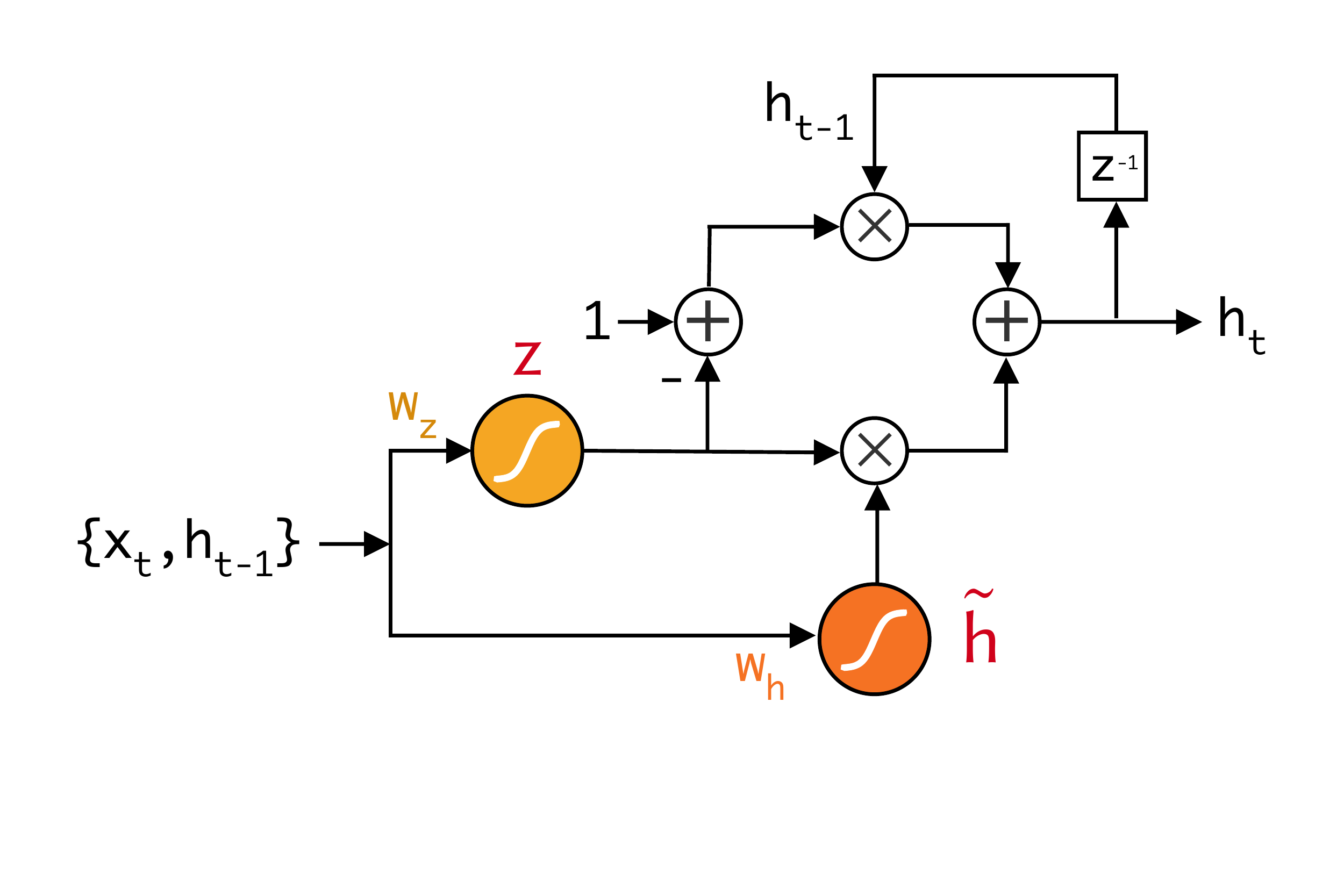}
        \caption{eGRU cell.}
        \label{fig:egrucell}
    \end{subfigure}
    \caption{ Circuit diagrams illustrating GRU (\ref{fig:grucell}) and eGRU (\ref{fig:egrucell}) cell architectures. Compared to GRU, eGRU omits the reset gate $r$ and does not require the weights, $w_r$. Additionally, it replaces sigmoid and $\tanh$ activation functions with softsign variants.}
    \label{fig:cellcircuit}
\end{figure}

\subsection{Activation Functions}
The original GRU cell utilized two kinds of activations; sigmoid functions ($\sigma$) in the update and reset gate equations, and the hyperbolic tangent function ($\tanh$) in the state update equation. However, since the M0+ lacks a dedicated floating point unit and DSP instruction set, executing either sigmoid or $\tanh$ functions are quite slow (see Table \ref{tab:activations}). For a more efficient recurrent unit, it was suggested in \cite{Ravanelli2017} that the $\tanh$ be replaced with a rectified linear unit (ReLU). Unfortunately, when combined with heavily quantized weights, recurrent cells with ReLU activations are too lossy to learn well. And since quantization is an essential part of our proposed eGRU architecture, ReLU activations are not an option.

A desirable alternative to the activation functions above is the softsign function. In \cite{glorot2010understanding}, softsign was shown to perform comparably with $\tanh$ and sigmoid in feed-forward networks. Although not as cheap (computationally) as ReLU, softsign is much cheaper than sigmoid or $\tanh$. A floating point implementation of the softsign function is more than 10x faster than either sigmoid or $\tanh$ functions on the M0+ (Table \ref{tab:activations}). Furthermore, the simplicity of the softsign function permits an even faster fixed point implementation. For these reasons, we adopt the softsign and a shifted version of it as activation functions for our proposed eGRU recurrent architecture.

\begin{table}
  \begin{subfigure}[b]{0.55\textwidth}
    \centering
    \begin{tabular}{|c|c|}
    \hline
    \textbf{GRU} & \textbf{eGRU} (this work)\\
    \hline
    $\begin{aligned}[t] 
        z_t &= \sigma(W_z \odot [h_{t-1},x_t]) \\
        r_t &= \sigma(W_r \odot [h_{t-1},x_t]) \\
        \tilde{h_t} &= \tanh(W_h \odot [ r_t * h_{t-1}, x_t]) \\
        h_t &= (1-z_t)*h_{t-1} + z_t*\tilde{h_t}
    \end{aligned}$ & 
    $\begin{aligned}[t] 
        z_t &= (\varsigma(W_z \odot [h_{t-1},x_t])+1)/ 2 \\ \\
        \tilde{h_t} &= \varsigma(W_h \odot [ h_{t-1}, x_t]) \\
        h_t &= (1-z_t)*h_{t-1} + z_t*\tilde{h_t}
    \end{aligned}$\\
    \hline
    \end{tabular}
    \caption{State equations of traditional GRU and proposed eGRU cells.}
    \label{tab:equations}
  \end{subfigure}\hfill
  \begin{subfigure}[b]{0.37\textwidth}
    {\renewcommand{\arraystretch}{1.18}
    \centering
    \begin{tabular}{|l|l|r|}
    \hline
    \multicolumn{1}{|c|}{Function} & \multicolumn{1}{c|}{Definition} & \multicolumn{1}{c|}{Latency} \\ \hline
    TanH                           & $f(x) = \tanh(x)$               & 328 us    \\
    Sigmoid                        & $f(x) = \frac{1}{1+e^{-x}}$     & 230 us    \\
    ReLU                           & $f(x) = \max(0,x)$              &  11 us    \\
    Softsign                       & $f(x) = \frac{x}{1 + |x|}$      &  24 us    \\ \hline
    \end{tabular}}
    \caption{Cost of activation functions on Arm M0+.}
    \label{tab:activations}
  \end{subfigure}

  \caption{\ref{tab:equations} Equations for GRU and eGRU highlights architectural differences between the two cells. In a single cell, gate weights $W_r$, $W_z$ are vectors of size $N+1$, where $N$ is the dimension of the input $x$. eGRU omits the reset gate ($r_t$) and uses softsign ($\varsigma$) activations instead of the $\tanh$ and sigmoid ($\sigma$) functions. \ref{tab:activations} Latency of standard activation functions executed on the M0+ MCU. Sigmoid and $\tanh$ functions are $10-30\times$ slower than ReLU and softsign functions. However, softsign is ideal since ReLU does not work well with extremely quantized weights.} 
  \vspace{-10pt}
  \label{tab:cellarchitecture}
\end{table}

\subsection{Weight Quantization}
Efforts to reduce neural network memory footprint often involve weight quantization. As networks are purely defined by learned parameters or weights, a reduction in memory for each weight through quantization results in tremendous shrinkage in the overall network size. Several studies have shown that neural networks are still effective even after weights are quantized to only 8-bits, leading to $20 - 49\times$ memory reduction \cite{Han2015,Wu2016}. Specifically in GRU architectures, ternarization (2-bit weights) has been proposed as feasible for recurrent neural networks at a small cost in performance \cite{Ott2016}.

However, we discovered that 2-bit quantization does not work well for our application. When combined with the single gate and activation function optimizations, ternarized weights in eGRU lead to poor performance. On the other hand, 3-bit quantization with septenary weights (7 levels) proved to be quite effective. Thus, 3-bit quantization was adopted for eGRU.

Besides the reduction in bits, our quantization scheme also ensures that quantized levels are negative integer exponents of two, similar to an approach in \cite{Ott2016}. This exponential quantization enables the replacement of weight multiplications with bit shifting, which in turn drastically reduces the computation time of an eGRU cell. The septenary weights used are: $[0,\pm0.25,\pm0.5,\pm1]$ and they are encoded using the mapping in Table \ref{tab:wtencoding} such that no external look-up tables are required at run-time. To multiply an input by a quantized weight, a simple, fast procedure featuring only bitwise operations is employed (Algorithm \ref{alg:qwtransform}).

\begin{table}
    \begin{minipage}{0.45\textwidth}
        \centering
        \begin{tabular}{|c|c|c|}
        \hline
        \textbf{Weight} & \textbf{Binary} & \textbf{Decimal} \\ \hline
        \textbf{+1.00}  & 000             & 0                \\
        \textbf{+0.50}  & 001             & 1                \\
        \textbf{+0.25}  & 010             & 2                \\
        \textbf{0}      & 111             & 7                \\
        \textbf{-0.25}  & 110             & 6                \\
        \textbf{-0.50}  & 101             & 5                \\
        \textbf{-1.0}   & 100             & 4                \\ \hline
        \end{tabular}
        \caption{Encoding of quantized weights. Encoding scheme is chosen for fast computation of weight transformations using only bitwise operations in Algorithm \ref{alg:qwtransform}. No additional look-up table is required at run-time.}
        \label{tab:wtencoding}
    \end{minipage} 
    \hfill
    \begin{minipage}{0.5\textwidth}
    \vspace{-25pt}
        \begin{algorithm}[H]
        \caption{Transformation by Quantized Weight}
        \label{alg:qwtransform}
        \KwIn{$x$, value to be scaled and, $w$, quantized weight.}
        \KwOut{Result of weight transformation, $y$.}
        $y \gets 0$\;
        \eIf {$w = 7$}{
           \Return $y$\;
           }{
           $y\gets x >>$ last 2 bits of $w$\;
           \lIf{bit $3$ of $w = 1$ }{$y\gets - y$}
           }
        \Return $y$
        \end{algorithm}
    \end{minipage}
\end{table}

\subsection{Fixed Point Arithmetic}
The final area of optimization in eGRU is the numeric format used for all math operations in the neural network. On better equipped processing units, single or double precision floating point formats are typically used. Unfortunately, since M0+ lacks an FPU, floating point operations are very costly (see Table \ref{tab:operations}). Hence, using solely integer operations in eGRU is desirable. 

Considering the M0+ has a 32-bit architecture, optimal execution of operations is achieved when operands are contained within 32-bits. Hence, we adopt the Q15 16-bit fixed point format for all arithmetic within eGRU. In Q15 format, 16-bit signed integers from 32,767 to 32,768 are used to represent decimals in the range of [-1,1) at intervals of $2^{-15}$. Necessary precautions ought to be taken during basic operations to prevent overflow and ensure closure under Q15. For instance, to divide two Q15 numbers, it is necessary to left-shift the dividend by 16 bits (yielding a 32 bit value) before undertaking the division to avoid losing precision. Hence, all eGRU operations (including activation functions) need to be translated to Q15 versions.

Weight multiplication is the most frequent operation in a neural network. Fortunately, by virtue of our exponential quantization, affine transformations for all layers in the network can be implemented by right-shift operations, which remains the same in Q15 format. The summation of all transformed inputs can exceed 16 bits and is thus accumulated in a 32 bit register. However, since the activation function is bounded by [-1,1), the output of the recurrent node remains a 16-bit Q15 number which can then be fed into yet another node. From simulations, we discovered that all inputs to an eGRU network will flow through the entire model in Q15 format and result in an output that is precise to at least 2 decimal places compared to those from an equivalent full precision (floating point) network.

An interesting modification worth mention pertains to the Q15 implementation of the softsign activation function. Inputs to activations, in the scheme described above, is a 32-bit accumulation of all transformed inputs. Since such inputs are already in 32-bit, undertaking a Q15 division would be impractical as it would require left-shifting, resulting in an overflow. One way to circumvent this is to clip the accumulated value to a certain domain to prevent overflow. We found that clipping to the domain (64,-64] resulted in a fast and accurate integer softsign approximation (see Table \ref{tab:operations}). A simple C++ definition of our softsign function is provided in Listing \ref{lst:softsign}.

\begin{table}
    \begin{minipage}[b]{0.50\textwidth}
    \centering
        \begin{tabular}{|c|c|c|}
        \hline
        \textbf{Operation} & \textbf{int32 (ns)} & \textbf{float32 (ns)} \\ \hline
        $\pm$              & 252                 & 4,853                 \\
        $<<,>>$            & 273                 & -                     \\
        $\times$           & 253                 & 6,578                 \\
        $\div$             & 1,923               & 12,942                \\ \hline
        softsign           & 4,740               & 24,910                \\ \hline
        \end{tabular}
        \caption{Comparison of time cost of integer vs floating point arithmetic on M0+. For basic maths, integer operations are beyond $10\times$ faster than floating point ones. Also, our fixed point approximation of the softsign is $5\times$ faster than the floating point implementation.}
        \label{tab:operations}
    \end{minipage}
    \hspace{10pt}
    \begin{minipage}[b]{0.40\textwidth}
    \vspace{-30pt}
    \centering
        \begin{lstlisting}[language=C++]
  int16 softsign(int32 x) {
      int32 a = x << 10;
      int32 b = (abs(x) + 32768) << 5; 
      int16 y = a / b;        
      return y;
  }
      \end{lstlisting}
    \captionof{lstlisting}{C++ routine for softsign function in Q15. Only integer operations are used. With this routine, softsign function is correctly defined for domain of (64.0,-64.0].}
    \label{lst:softsign}
    \end{minipage}
    \vspace{-15pt}
\end{table}

\subsection{Summary of Optimizations}
A summary of the proposed optimizations and their corresponding reductions in computational resource requirements is provided in Table \ref{tab:optsummary}. Table \ref{tab:cellrequirements} presents a breakdown of the resource requirements for an eGRU cell is compared with that for a traditional GRU cell. The overall number of operations are fewer for eGRU (12) than for GRU (17). Furthermore, eGRU trades off most multiplications for cheaper and faster bit shift operations. Also, note that all eGRU operations are integer operations (INOPs) compared to the floating point operations (FLOPs) in GRU. 

Memory and latency reduction scale factors for each optimization is also listed in Table \ref{tab:cellrequirements}. Quantization mostly accounts for the miniaturization, reducing the cell size by a factor of 10. On the other hand, using solely Q15 integer operations rather than floating point operations drastically speeds up cell execution and yields a 20$\times$ reduction in latency. Analytically, softsign activations also account for significant reduction in latency (10$\times$), even without accounting for the additional gains in speed afforded by the faster integer softsign approximation in Listing \ref{lst:softsign}.

\begin{table}
\begin{subfigure}[t]{0.36\textwidth}
    \centering
    \begin{tabular}{|l|c|c|c|}
    \hline
    \textbf{Requirement} & \textbf{eGRU} & \textbf{GRU} \\ \hline
    \# of Parameters      & 6                 & 9         \\    
    Additions           & 6                 & 8         \\
    Multiplications     & 2                 & 9         \\
    Bit Shifts          & 4                 & -         \\ \hline
    \end{tabular}
    \caption{Comparison of cell requirements.}
    \label{tab:cellrequirements}
\end{subfigure}\hspace{1pt}
\begin{subfigure}[t]{0.62\textwidth}
    \centering
    \begin{tabular}{|l|c|c|c|}
    \hline
    \textbf{Optimization} & \textbf{Memory} & \textbf{Latency} & \textbf{Modification}                     \\ \hline
    Single Gate Unit      & 1.5$\times$     & 1.5$\times$             & $\,\,$3 gates $\Rightarrow$ 2 gates              \\
    Quantization          & $\,$10$\times$      & -                & 32 bits $\Rightarrow$  3 bits              \\
    Softsign Activations  & -               & $\,$10$\times$              & $\sigma$ \& $\tanh$ $\Rightarrow$ softsign \\
    Q15 Arithmetic        & -               & $\,$20$\times$              & $\,\,$FLOPs $\Rightarrow$ INOPs                  \\ \hline
    \end{tabular}
    \caption{Reduction scale factors for individual optimizations.}
    \label{tab:reductions}
\end{subfigure}
    \caption{Summary of computational requirements and savings of an eGRU cell compared to GRU. \ref{tab:cellrequirements} Comparison of number of parameters and operations required for each cell type. eGRU's fewer parameters decreases required memory storage and its fewer operations reduces execution latency. \ref{tab:reductions} Tabulation of memory and latency reductions afforded by proposed optimizations. Quantization achieves the most memory reduction by a factor of 20, whereas Q15 integer implementation attains the most speed up at a 10$\times$ reduction in latency.}
    \label{tab:optsummary}
\end{table}

\subsection{Training Setup}
Since the sole interest is inference (not learning) on target devices, training of eGRU cells can be undertaken on high-end computers with enormous resources,  using cutting-edge deep learning libraries like PyTorch \cite{paszke2017pytorch} and Tensorflow \cite{abadi2016tensorflow}. However, due to the modifications proposed above, additional steps are still necessary to effectively train eGRU cells in a neural network. Specifically, the effects of quantization and fixed point arithmetic in eGRU need to be simulated during the training process as well. 

One common approach to quantize neural network weights is by means of post-training step. The network is first trained with full-precision weights (32- or 64-bit floating point numbers), then learned weights are quantized to the desired precision. This approach requires no adjustments to the training process since quantization is applied afterwards. While this method is suitable for reducing model sizes, it is not robust especially with heavy quantization like ours. That is because during training, the network is oblivious to the post-training quantization and therefore cannot learn to accommodate it.

A more desirable approach is to implement the quantization process as part of the network's computational graph so it can be incorporated into the training process. Yet, since quantization is discontinuous, it cannot be incorporated in backpropagation training without adjustments. To address this, we adopt an approach similar to that used by \citet{Hubara2016}. During forward pass of backpropagation, weights are first cached and passed through a quantization function before they are used in the dot product. Then during backward pass, gradients are computed with respect to the stored full-precision weights.

\begin{figure}[t]
    \begin{subfigure}[b]{0.36\textwidth}
        \includegraphics[width=\textwidth]{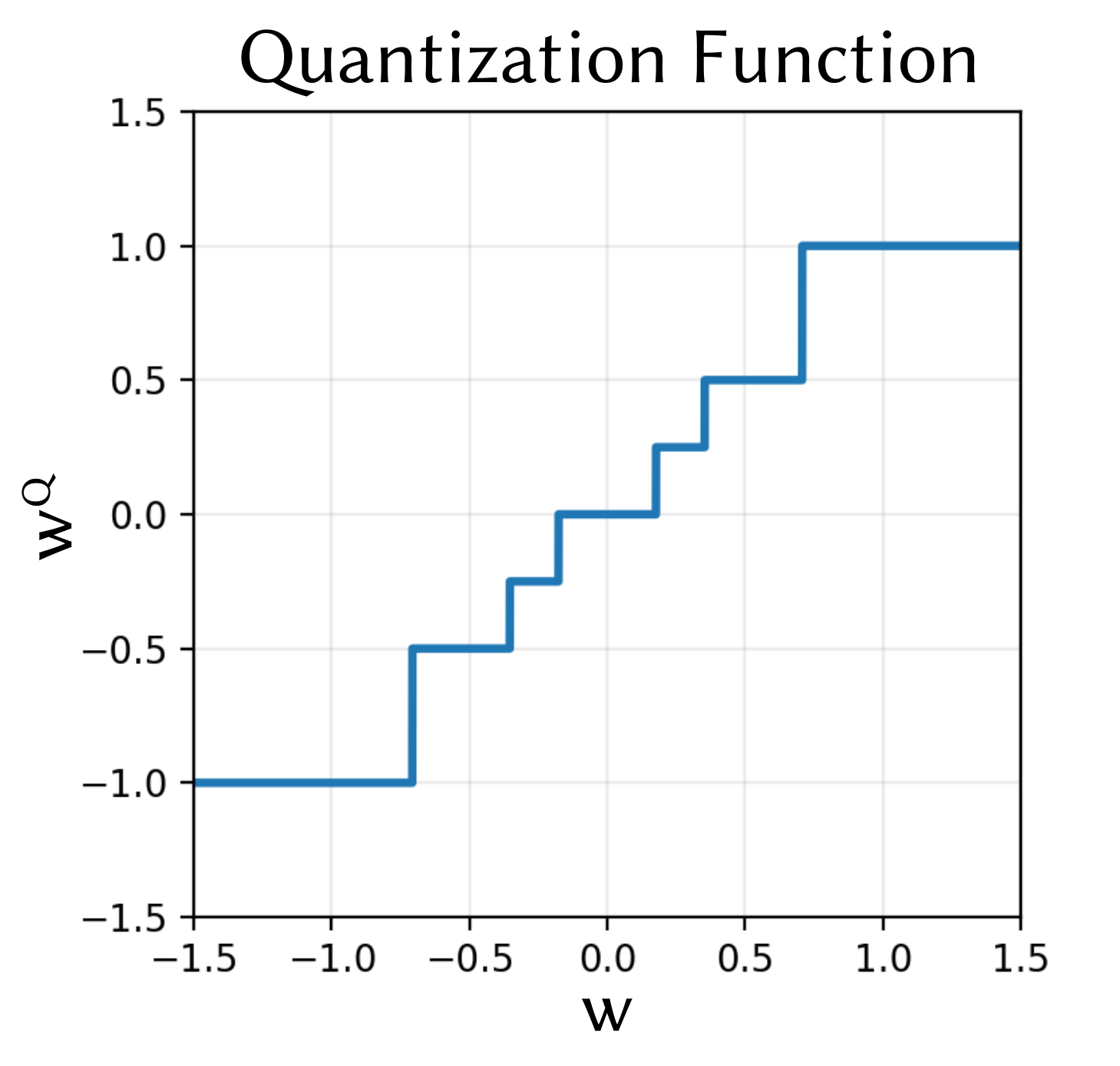}
        \caption{Exponential quantization function.}
        \label{fig:quantization}
    \end{subfigure}\hfill
    \begin{subfigure}[b]{0.60\textwidth}
        \includegraphics[width=\textwidth]{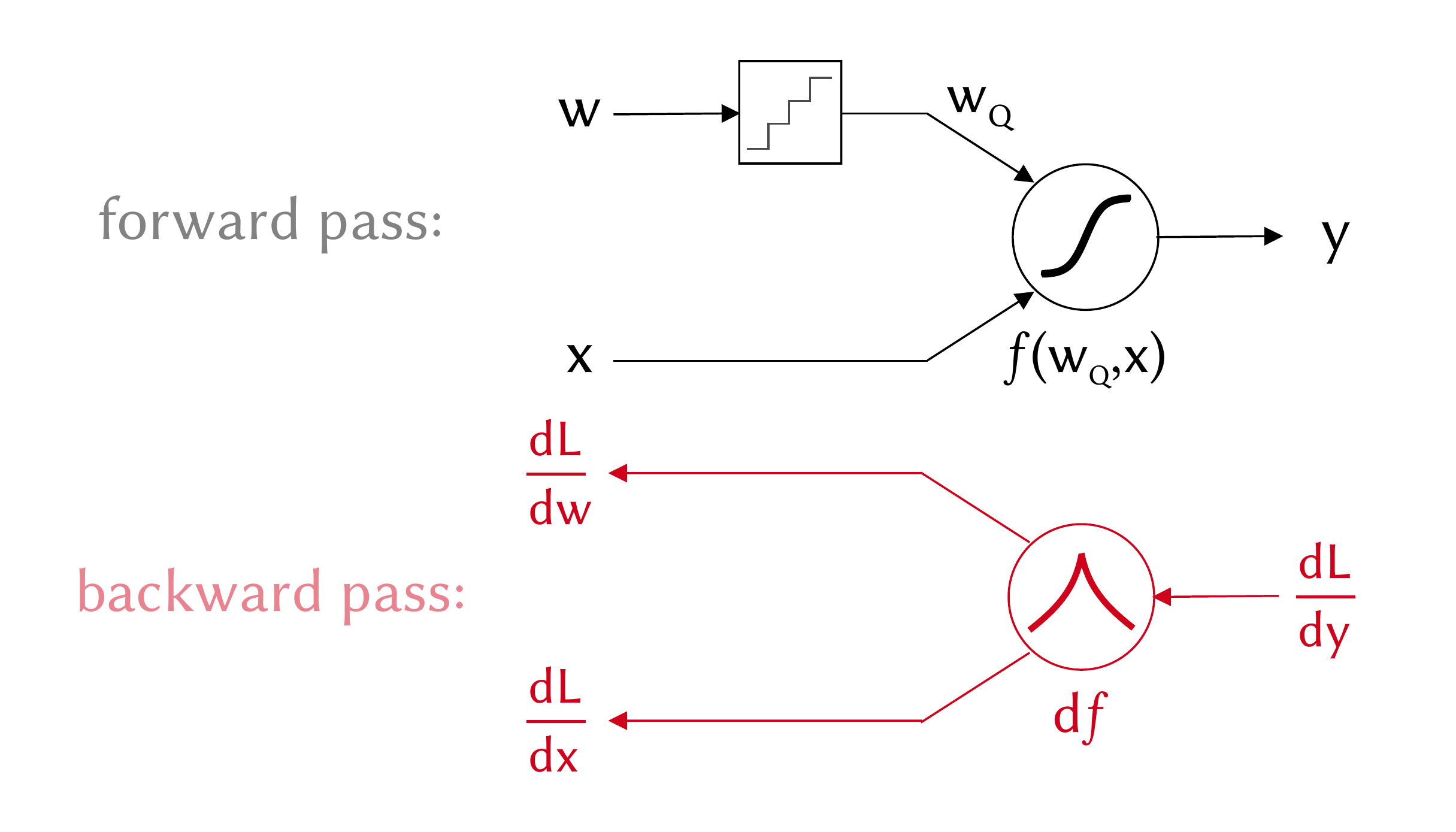}
        \caption{Quantization in backpropagation.}
        \label{fig:backprop}
    \end{subfigure}
    \caption{Modifications to include quantization in training process. \ref{fig:quantization} Quantization function for yielding septenary weights that are negative integer exponents of 2. \ref{fig:backprop} During forward pass, quantization is applied to weights before they are used in any transformation. In backward pass, gradients are computed with respect to stored full precision weights.}
    \label{fig:trainingsetup}
\end{figure}

An illustration for this is provided in Fig. \ref{fig:trainingsetup} and the quantization function used is given by: 
\[ W_Q = \begin{cases} 
      +1, & W \geq +1 \\
      \;\;\, 0,  &\!\!\! |W| \leq 0.25\\
      -1, & W \leq -1 \\
      \text{sgn}(W) \cdot 2^{\nint{(\text{log}_2|W|)}} & \text{otherwise}  
   \end{cases}
\]
where $W$ is the full-precision weight, and $W_Q$ is the quantized weight. In a sense, quantization is realized just like weight clipping (where weights are restricted to a defined range), and can thus be similarly considered a regularization \cite{merolla2016deep}. Note though, that once the network is trained, full-precision weights are discarded and only quantized weights are used in inferencing. 

Besides quantization, simulating eGRU's fixed point arithmetic also requires training adjustments. The Q15 fixed point format used can only represent decimal numbers in the range (1,-1]. Therefore, during training, all network operations have to be constrained to fit within these Q15 bounds, in order to avoid overflow on the target device. In our work, we achieve this by clipping all dot products and activations to the (1,-1] range. The weight clipping here, as alluded to above, serves as yet another regularizer for eGRU networks.

\section{Experimentation}
To evaluate the proposed features of eGRU, four experiments are undertaken. The first experiment investigates the benefits of our architectural choices, mainly the single gate mechanism and the activation functions. The second experiment analyzes the isolated impact of the various optimizations on the performance of a recurrent network. The third experiment benchmarks the computational speed of an eGRU cell on an M0+ in contrast to the GRU and RNN. Finally, the last experiment evaluates the performance of a fully optimized eGRU neural network embedded onto the target low power device for acoustic event detection tasks. The tasks, network architecture and experimental setups are discussed below in detail.

\subsection{Task Description}
To evaluate the various recurrent cells, three Acoustic Event Detection (AED) tasks of varying complexity are considered. They are as follows:

\begin{itemize}[itemsep=4pt,topsep=6pt]
    \item \textbf{Cough Detection:} The first and least complex of the three task is the classification of short acoustic events as cough, speech or other non-cough sounds. Other non-cough sounds consists of voiced non-speech sounds, ambient sounds and noise (white and pink). This task is actually the application of interest for our wearable device. Since there are only three possible classes, the problem is much easier than the other tasks. Furthermore, because cough sounds have a relatively short duration (350 ms on average), this problem would not require that much long-term memory compared to the others. The dataset used here is from a prior work \cite{amoh2016deep}, and consists of 2,518 audio sounds. The dataset is not exactly balanced and contains more speech and other sounds than actual cough events.

    \item \textbf{Spoken Digits:} The next task involves distinguishing between ten spoken digits (numbers 0 to 9). The audio data is obtained from Google's Speech Commands Dataset \cite{Warden2018} which comprises of audio recordings covering 20 simple commands. In this work, we use a subset of that dataset, consisting of 23,666 one-second audio clips. There are ten digits and the dataset is well balanced, with roughly 2,400 examples per digit. On one hand, this task can be considered more complex than the cough detection problem because there more classes and audio samples are roughly 3$\times$ longer. On the other hand, this dataset is 10$\times$ larger and thus should be easier to learn.  

    \item \textbf{Urban Sounds:} The third and most challenging of the tasks considered involves identifying different urban sounds. The UrbanSound8k dataset is used \cite{salamon2014dataset} and it includes 8,732 sounds recorded from 10 urban environments. Thus, there are ten different classes in this problem as well, and some examples are: \textit{children\_playing}, \textit{dog\_barking}, \textit{car\_horn}, \textit{siren} and \textit{street\_music}. Certain classes sound very similar and it can be difficult for even a human to accurately identify some entries. Moreover, entries have a maximum duration of 4 seconds, which is much longer than the other two tasks and would therefore require larger long-term memory capability.
\end{itemize}

\begin{figure}
    \begin{subfigure}[b]{0.60\textwidth}
    \centering
        \includegraphics[width=\textwidth]{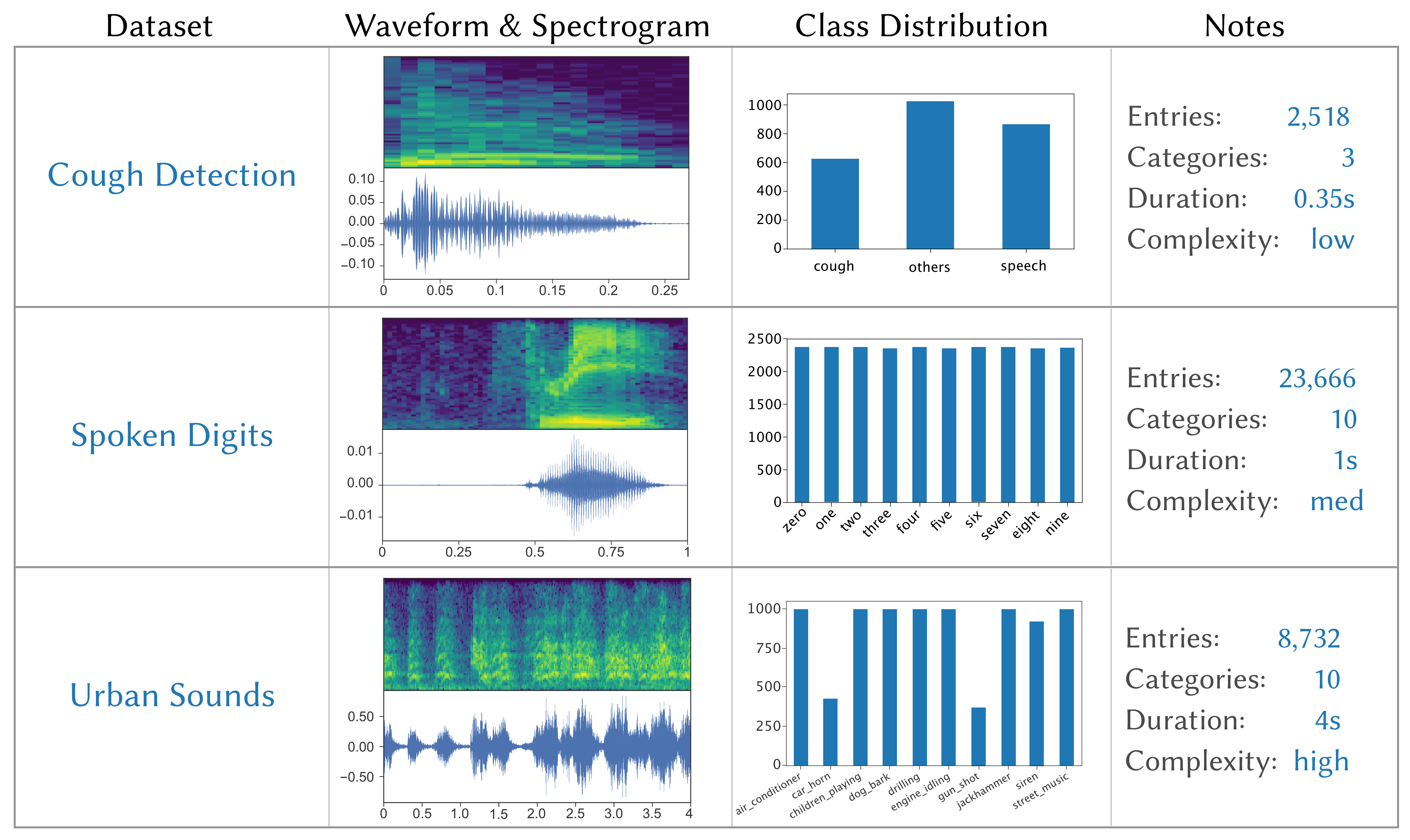} 
        \caption{Dataset and tasks for training neural networks.}
        \label{fig:aed_datasets}  
    \end{subfigure}
    \hfill
    \begin{subfigure}[b]{0.38\textwidth}
    \centering
        \includegraphics[width=\textwidth]{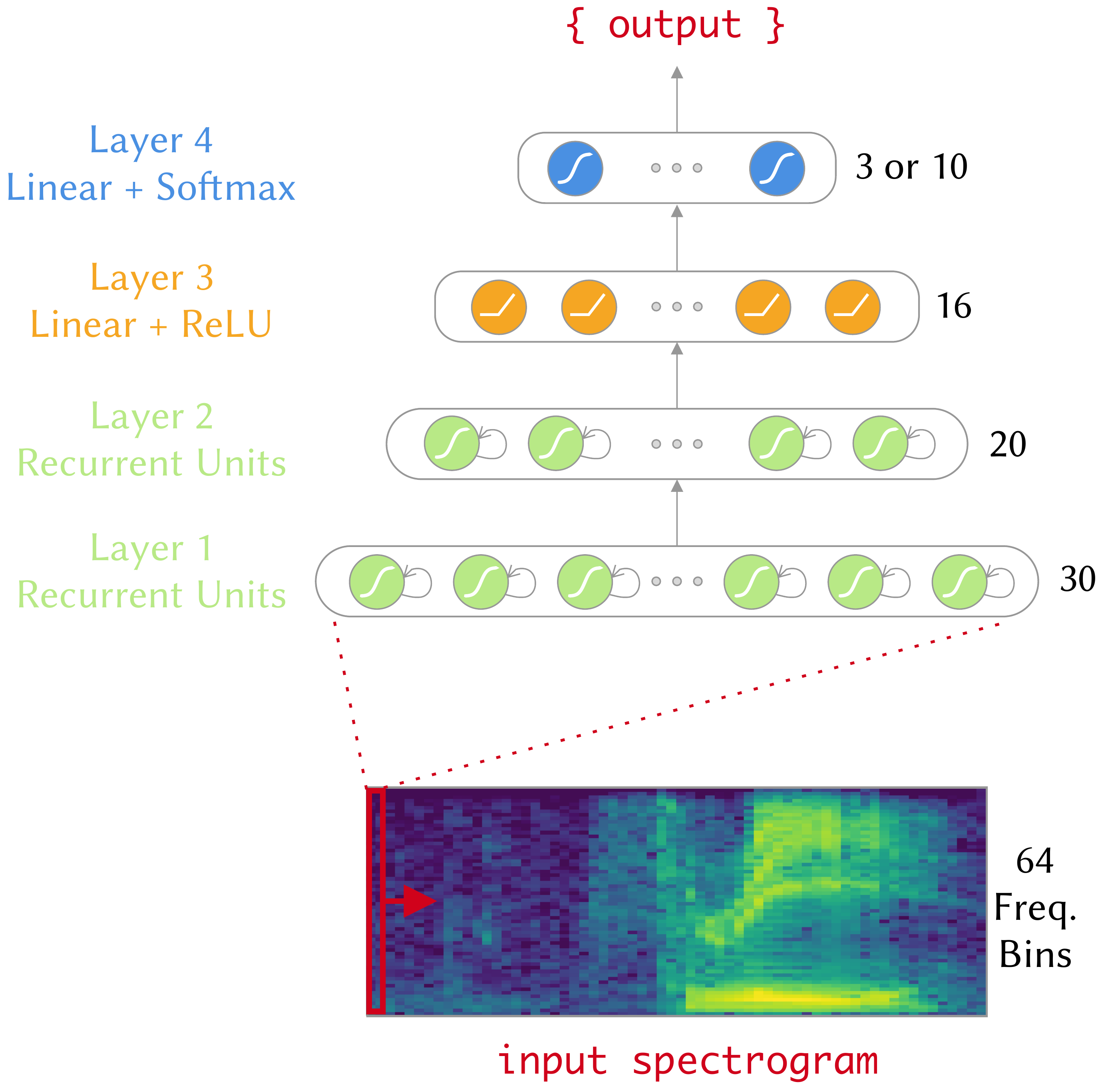}
        \caption{Recurrent neural network architecture.}
        \label{fig:network}  
    \end{subfigure}
    \caption{\ref{fig:aed_datasets} Summary of the three audio event detection (AED) datasets and tasks investigated in experiments. Cough detection is easiest because samples are relatively short and it is a 3-class problem. Urban sounds are 10$\times$ longer and consist of 10 different classes. \ref{fig:network} An illustration of the neural network architecture used for all tasks. Recurrent units in the first two layers are RNN, eGRU or GRU cells. The input is a sequence of FFT vectors.}         
\end{figure}

A summary of the three tasks is provided in Fig. \ref{fig:aed_datasets}. Audio from all datasets are pre-processed in a similar manner. First, recordings are downsampled from their original sampling rates to 8 KHz, then further processed into spectrograms using the Short-Time Fourier Transform (STFT). For the STFT, a window length of 128 samples is used, with no zero-padding or overlap. These analysis configuration are motivated by what can feasibly be computed on the M0+ MCU in a real-time application. For each entry in the corresponding dataset, the spectral analysis yields a 64$\times$24 spectral chunk (64 frequency bins and 24 time steps) for the Cough task, 64$\times$64 for Spoken Digits and 64$\times$250 for Urban Sounds.

\subsection{Network Architecture}
For all three tasks, the same recurrent neural network architecture is used. Though better performing architectures can be used for each task, fixing the architecture makes for a fair comparison across tasks and experiments. Also, it is also common practice to use similar network architectures when introducing new recurrent units to enable direct comparison of architectural benefits \cite{chung2014empirical,collins2016capacity}. In our case, we are primarily interested in the differences in performance for a drop-in replacement of all GRU cells with eGRU. Thus, it is preferred to keep the same network architecture for both eGRU and GRU cells. A four layer network is used, featuring two recurrent layers with 30 and 20 corresponding recurrent units (eGRU, GRU, or RNN cells). This is preceded by a 16 node linear layer with ReLU activations. The fourth and final layer is the classification layer with task specific number of output nodes and preceded by a softmax function. An illustration of this neural network architecture is provided in Fig. \ref{fig:network}.

The neural networks are implemented and trained using PyTorch on a computer with 64GB RAM, a 3.8 GHz Core i5 CPU and an Nvidia GTX 1080 Ti GPU. The ADAM optimizer is used, with mini-batch sizes of 128. Training is undertaken for a maximum of 200 epochs with checkpointing at each epoch to retain only the best performing model state. In checkpointing, if the network's loss on a validation set is the minimum seen so far, the network's state is saved as the current best model. Model checkpointing therefore enables training for much longer epochs without the risk of overfitting and ensures the best model state is used eventually for evaluation.

\subsection{Experiment 1}
In the first experiment, we seek to answer the following question: how does the proposed eGRU architecture compare with traditional recurrent architectures on acoustic event detection tasks? Of particular interest here is the performance of the single gate mechanism together with the softsign activation functions because those are distinctive features of the eGRU. Quantization and integer arithmetic, on the other hand, can be applied to any recurrent cell and are therefore not considered in this experiment.

To evaluate eGRU's architecture, RNN, GRU and eGRU networks are trained on the three datasets using the network architecture in Fig. \ref{fig:network}. While eGRU is a single-gate version of the GRU, the vanilla RNN is in a sense an even simpler, ``zero-gate'' GRU. We include the RNN in this experiment to explore whether gated models like the GRU or eGRU are even necessary for our application. A 10-fold cross-validation scheme is adopted for a thorough evaluation. On each task, classification accuracies on test sets are computed for the different cell types and aggregated across all folds. In all, it takes about 5 days to train all models across the ten folds.

\subsection{Experiment 2}
Next, the second experiment explores how the proposed optimizations, in isolation, impact performance of a GRU network. To that end, four variants of the network architecture in \ref{fig:network} are implemented featuring: (1) full precision original GRU cells, (2) single gated recurrent cells (3) 3-bit quantized weights in GRU cells and linear layers and (4) softsign activations in GRU cells. These networks are to illustrate the performance costs of individual optimizations. However, integer arithmetic is not included in this analysis because it mainly affects run-time speed on the embedded system and has little effect on network performance in terms of accuracy.

All four networks are trained on the above mentioned AED datasets. Also, a 70\% : 30\% training and test set split is used rather than the 10-fold cross-validation scheme used in the first experiment. From the 70\% subset of the dataset used for training, 25\% is further set aside as the validation set to determine when training converges.

\subsection{Experiment 3}
The third experiment aims to measure the latency of the each recurrent cell type on the Arm M0+ MCU. Measuring latency will reveal and quantify the gains in speed and memory that eGRU cell has over the GRU cell. Consequently, single eGRU and GRU cells are implemented and benchmarked on the target M0+ device. The eGRU cell here features all optimizations proposed including quantization and fixed point arithmetic. 

For each recurrent cell type, a sequence of 1000 one dimensional inputs ranging from -1.0 to +1.0 are iteratively fed to the cell. The time taken for each cell to process the sequence on the M0+ is measured and reported as the latency benchmark. Note that while the GRU cell and its activations are implemented in 32-bit floating point format, eGRU is implemented solely in 16-bit integer data types. The latency benchmarks provide a concrete measure of how much faster and smaller eGRU is compared to GRU on an M0+ MCU.

\subsection{Experiment 4}
In the fourth and final experiment, the goal is to evaluate the performance of eGRU models trained on the three datasets and deployed to the M0+ for inference. For each dataset, a fully optimized eGRU model with the same architecture as in Fig. \ref{fig:aed_datasets} is trained. Since deployment is the aim, eGRU cells used here feature all optimizations: single gate architecture, 3-bit quantization, softsign activations and a Q15 implementation. The same 70\% : 30\% cross-validation scheme used in experiment 2 is also adopted here. 
 
Once the network is trained, quantized network weights are included in the firmware of the target M0+ device. At test time, test audio samples are fed to the M0+ device via a USB serial communication for processing and classification. The entire audio classification by the eGRU model takes place on the device, which then returns a prediction afterwards. Predictions are aggregated and analyzed to obtain performance metrics. Also, the memory, computations and speed of the embedded model are analyzed. For comparison, an equivalent GRU network is implemented on a workstation computer and evaluated using the same train and test sets.

\begin{figure}[b]
    \centering

    \begin{minipage}[b]{0.58\textwidth}
    \centering
    {\renewcommand{\arraystretch}{1.3}
    \begin{tabular}{|l|c|c|c|}
    \hline
    \multicolumn{1}{|c|}{\textbf{Cell}} & \textbf{CoughDetect}   & \textbf{SpokenDigits}  & \textbf{UrbanSounds}     \\ \hline
    RNN  & 88.0 $\pm$ 2.1  & 34.2 $\pm$ 5.7  & 51.0 $\pm$ 2.4   \\
    eGRU & \textbf{92.1 $\pm$ 1.7}  & \textbf{92.3 $\pm$ 0.5}  & \textbf{74.3 $\pm$ 2.1}  \\
    GRU  & \textbf{92.5 $\pm$ 1.9}  & \textbf{93.1 $\pm$ 0.6}  & \textbf{76.2 $\pm$ 2.7}  \\ \hline
    \end{tabular}}
    \captionof{table}{Experiment 1 results reporting average classification accuracies (\%) for the recurrent cells across 10-folds in three AED tasks. The eGRU architecture performs comparably with GRU across all tasks.}
    \label{tab:expt1res}
    \end{minipage} \hfill
    \begin{minipage}[b]{0.4\textwidth}
    \centering
        \includegraphics[width=\textwidth]{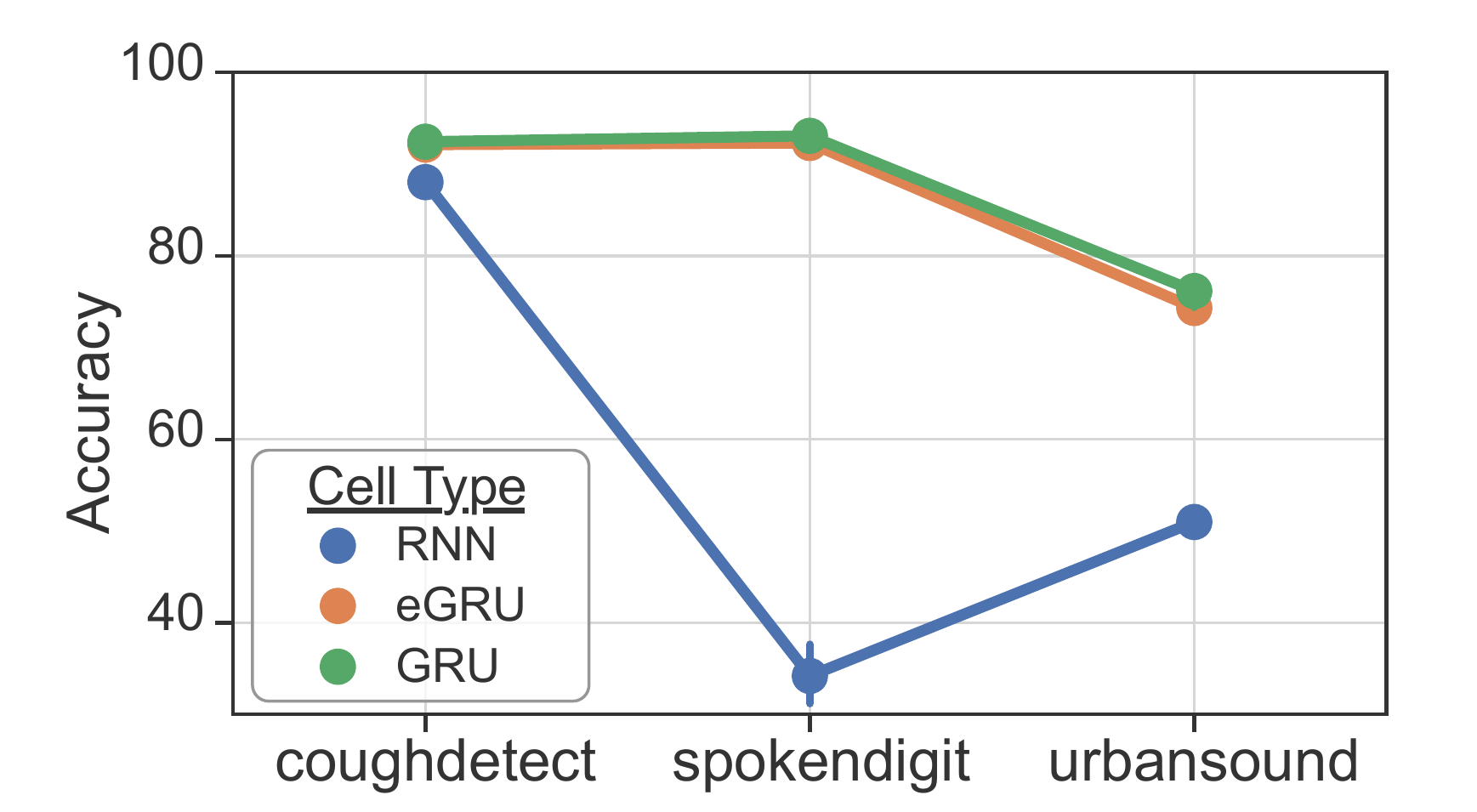}
        \vspace{-20pt}
        \captionof{figure}{Plot visualizing accuracies from Table \ref{tab:expt1res}. Tasks have varying complexities. Specialized cells perform better than RNN.}
        \label{fig:expt1res}
    \end{minipage}    
\end{figure}

\section{Results}
\subsection{Experiment 1}
Table \ref{tab:expt1res} and Fig. \ref{fig:expt1res} report the classification accuracies across 10-folds for the various recurrent cells in the first experiment. As expected, performance in general drops as the sequence length and task complexity increases across tasks. Cough detection, with a short sequence length of 24, is shown to be the easiest task where even RNN does well. On the other hand, in urban sound identification where sequences are 10$\times$ longer, even the gold standard GRU suffers a notable decrease in performance. 

\begin{figure}[t]
    \vspace{-5pt}
    \includegraphics[width=\textwidth]{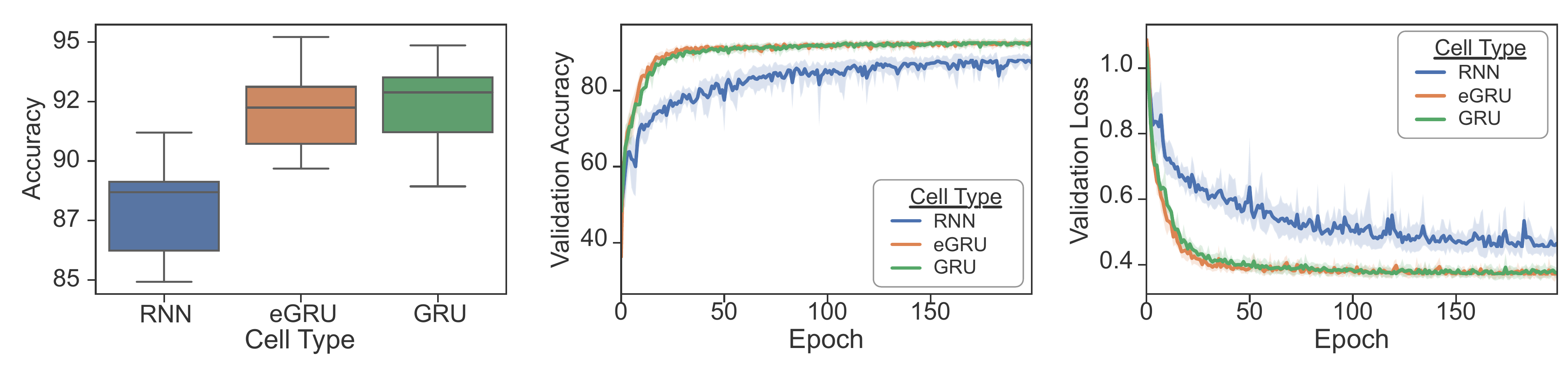}
    \caption{Cough detection task performance and training curves.}
    \label{fig:coughdetect}            

    \includegraphics[width=\textwidth]{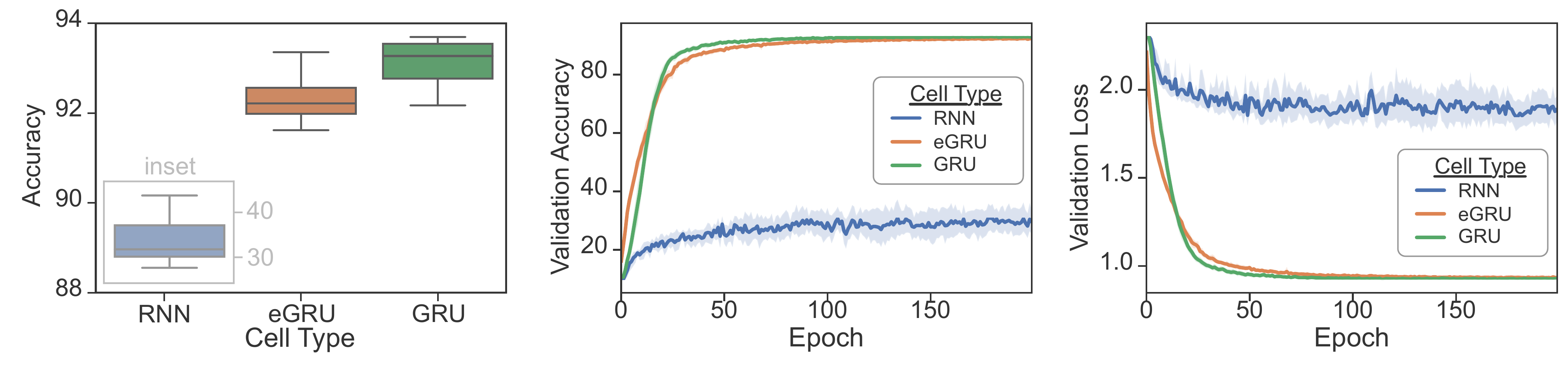}
    \caption{Spoken digits task performance and training curves.}
    \label{fig:spokendigit}            

    \includegraphics[width=\textwidth]{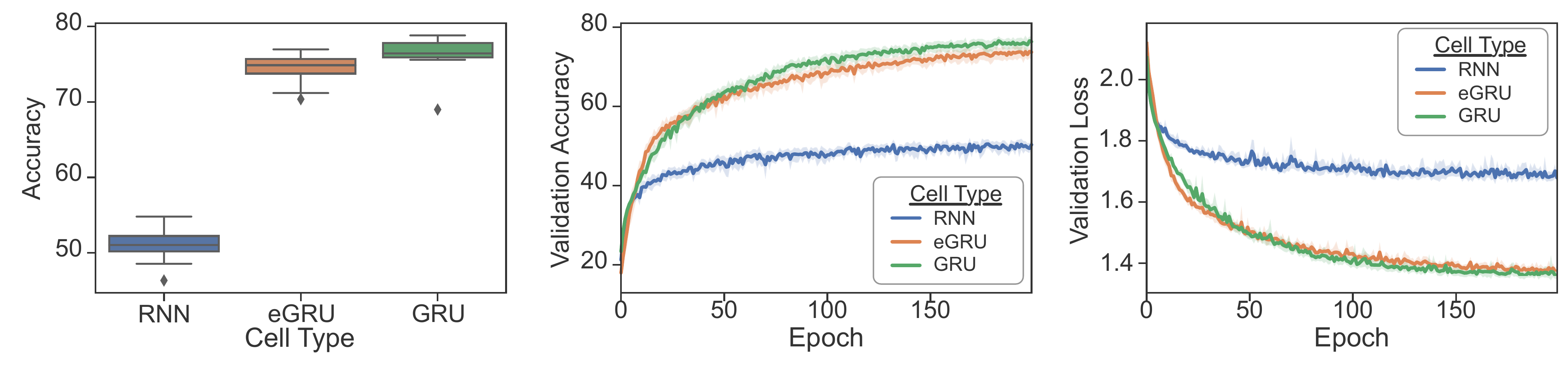}
    \caption{Urban sounds task performance and training curves.}
    \label{fig:urbansound}              
\end{figure}

In all tasks, the baseline RNN cell significantly under-performs both GRU and eGRU cells, especially in the more complex tasks. This trend is observed in the accuracy scores in Table \ref{tab:expt1res} and in the per-dataset box-plots and validation curves in Fig. \ref{fig:coughdetect}-\ref{fig:urbansound}. Counter to expectations, RNN actually performed worse on spoken digits classification rather than the longer, seemingly more complex urban sounds identification. A possible explanation for this observation is that although audio entries from the spoken digit dataset are one second long, the actual event contained is often much shorter than that as evident in Fig. \ref{fig:aed_datasets}. Since recordings begin and end with long periods of silence or background sounds, traditional RNN cells lacking advanced memory capacity will struggle to identify the buried event. The validation curves in Fig. \ref{fig:spokendigit} attests to this, showing the RNN cells' inability to learn much in spoken digits task. Such behavior is not observed in the more advanced GRU and eGRU cells. In all, the under-performance of RNN cells confirms that gated recurrent architectures like eGRU and GRU are indeed beneficial for the AED tasks in question. 

The eGRU model performs consistently well across all tasks, closely following GRU's performance. Even in the difficult urban sounds task, eGRU only suffers a 2\% loss in accuracy relative to GRU. The same trend is observed in the training curves in Fig. \ref{fig:coughdetect}-\ref{fig:urbansound}, where eGRU follows closely behind GRU. That eGRU performs decently even on longer sequences is interesting because one would expect the lack of a reset gate to greatly inhibit the cell's long-term memory. However, it turns out that with full precision weights, eGRU can indeed accomodate longer sequences.

From the observations mentioned above, the proposed eGRU architecture, mainly the single gating mechanism, appears to be suitable for AED tasks. eGRU is certainly preferable to RNN and a good alternative to GRU in keyword-spotting applications.

\subsection{Experiment 2}
\begin{figure}
    \begin{subfigure}[b]{0.58\textwidth}
    \centering
        \includegraphics[width=\textwidth]{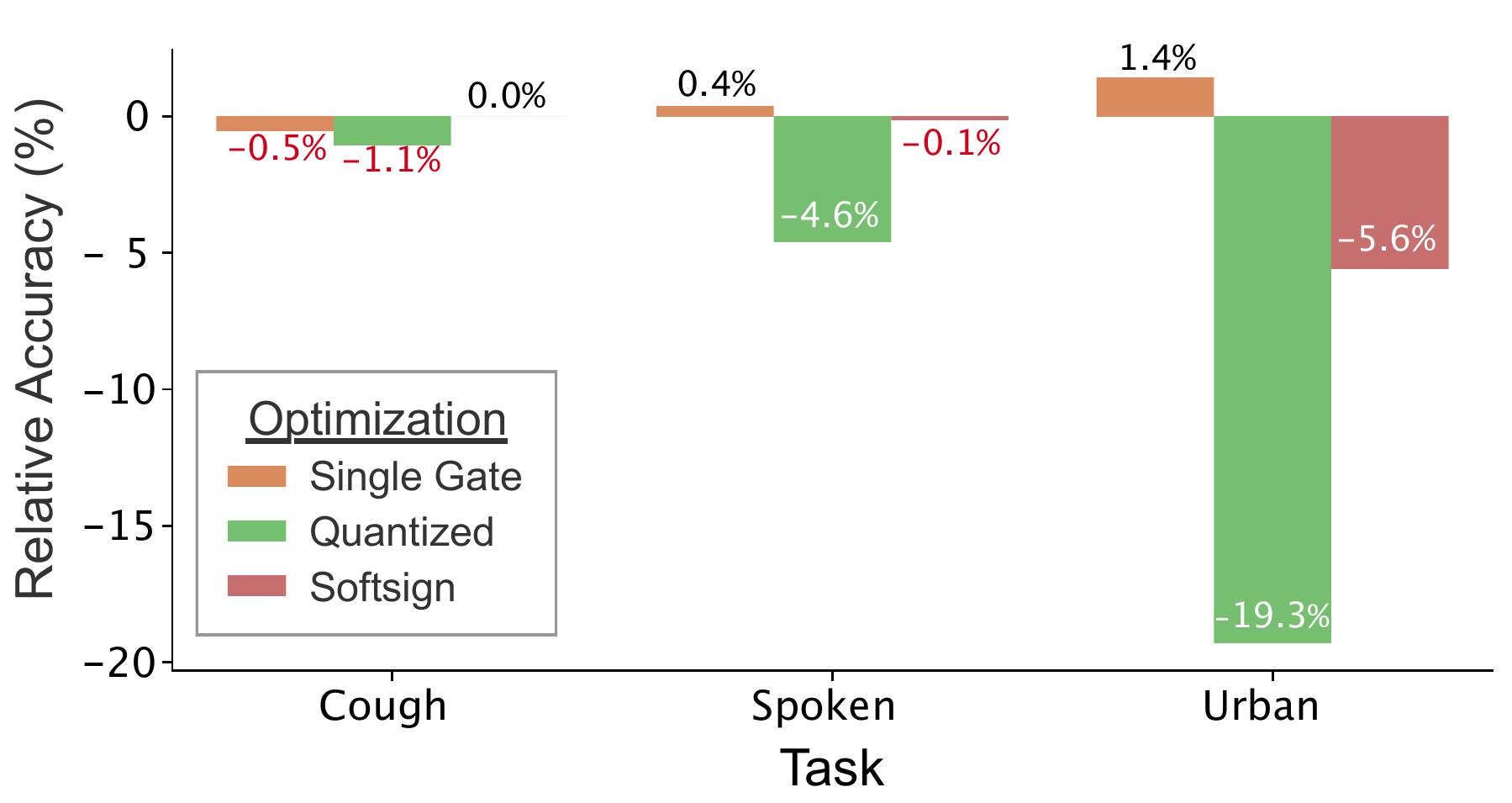} 
        \caption{Bar chart comparing performance of individual optimizations.}
        \label{fig:ablation_barplot}  
    \end{subfigure}
    \hfill
    \begin{subfigure}[b]{0.40\textwidth}
    \centering
        \includegraphics[width=\textwidth]{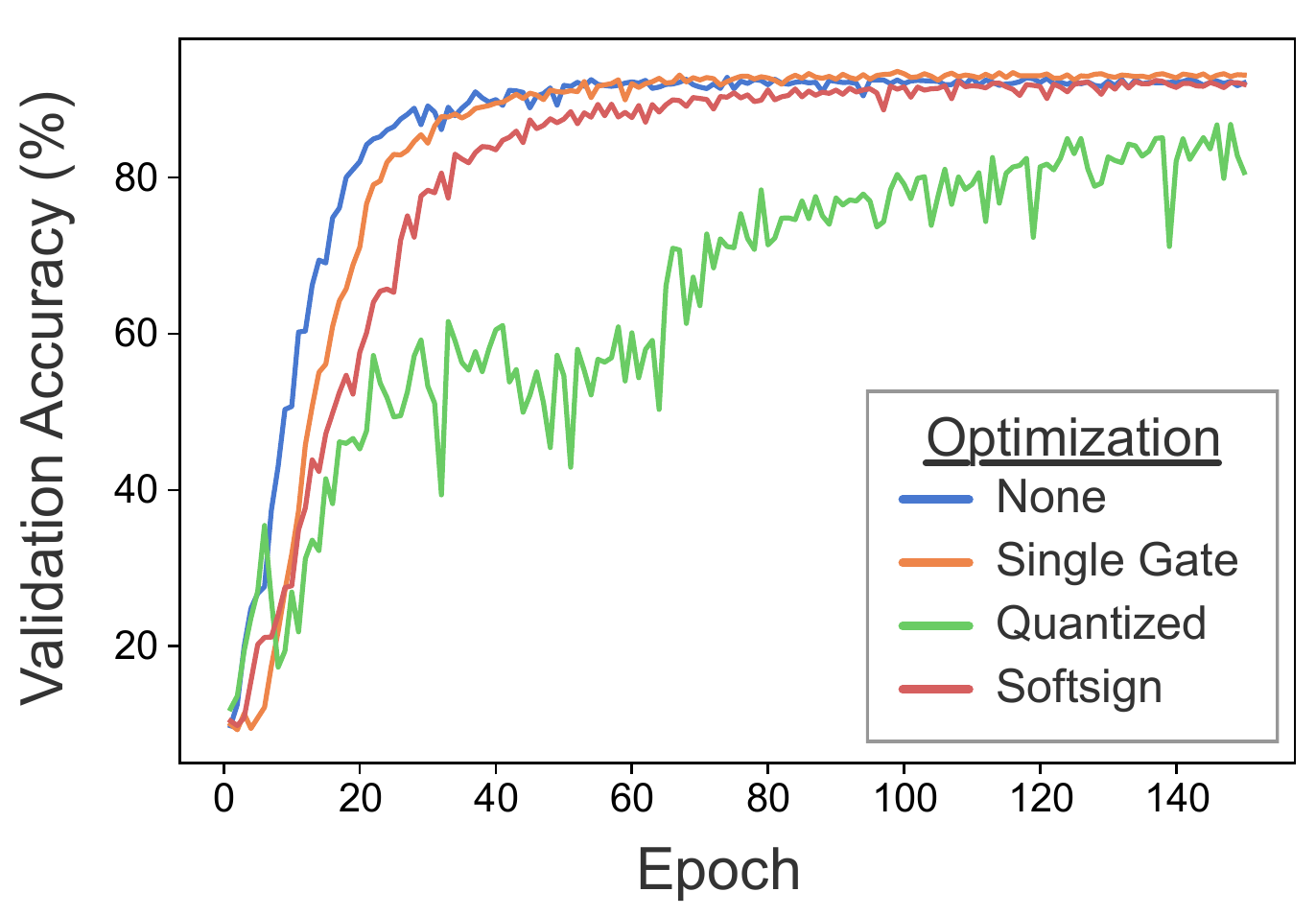}
        \caption{Validation curves for spoken digit task.}
        \label{fig:ablation_curves}  
    \end{subfigure}
    \caption{Experiment 2 results analyzing the isolated impacts of proposed eGRU optimizations on performance. \ref{fig:ablation_barplot} Relative classification accuracies for optimizations with respect to original GRU, across three AED tasks. 3-bit quantization affects accuracy the most, resulting in up to 19\% reduction in accuracy. \ref{fig:ablation_curves} Corresponding training curves for different optimizations in spoken digit task. All proposed optimizations make networks harder to training, requiring more epochs to converge compared to original GRU.}  
    \label{fig:ablation}       
\end{figure}

The ablation study results are presented in Fig. \ref{fig:ablation}. In Fig. \ref{fig:ablation_barplot}, classification accuracies for isolated optimizations are compared to that of a baseline GRU without any optimization. In all three AED tasks, 3-bit quantization impacts network performance the most, resulting in 1-20\% reduction in accuracy. The effects of the single gate and softsign activations are not as pronounced except for the harder urban sounds task. In the longer urban sounds task, softsign activations evidently hurt performance (by $\sim$6\%) whereas single gate architecture appears to slightly help. 

The validation curves on the spoken digits tasks provided in \ref{fig:ablation_curves} also reveal how the individual optimizations influence the network training. From the curves, all optimizations are seen to make training more difficult, taking longer time to converge than the original GRU. Once again, 3-bit weight quantization has the most adverse effect on training, taking much longer to converge and exhibiting noisy performance due to the extremely low precision weights. Softsign activations also makes it harder to train, requiring 20 more epochs to catch up to the original GRU in accuracy. The single gate optimization impacts training the least.

\subsection{Experiment 3}
The results for the third experiments are provided in Table \ref{tab:expt3res} and it informs on the computational costs of the different recurrent cells. Evidently, eGRU's use of solely integer operations result in much faster execution on the M0+ compared to GRU. On the M0+, an eGRU cell takes only 13 ns to execute, which is more than 60$\times$ faster than GRU. Besides speed, the extreme 3-bit quantization of eGRU results in a 12$\times$ reduction in memory compared to GRU (with 32-bit floats). A single eGRU cell is only 3 bytes large. 

\begin{table}
    \centering
    \begin{tabular}{|c|r|r|r|}
    \hline
    \textbf{Cell} & \textbf{Operations} & \textbf{Memory} & {\textbf{Latency}} \\ \hline
    eGRU  & 12 INOPs    &  \textbf{3 bytes} &\textbf{13 ns}   \\
    GRU   & 17 FLOPs    & 36 bytes          & 794 ns          \\ \hline
    \end{tabular}
    \caption{Experiment 3 results showing benchmark of M0+ latency for eGRU versus GRU cells. An eGRU cell requires only integer operations (INOPs) instead of the more expensive floating point operations (FLOPs) in traditional GRU. On the M0+, eGRU is \textbf{60$\times$ faster} and \textbf{12$\times$ smaller} than GRU.}
    \label{tab:expt3res}
\end{table}

\subsection{Experiment 4}
\begin{figure}
    \begin{subfigure}[b]{0.38\textwidth}
    \centering

    \begin{tabular}{|l|c|c|}
    \hline
    \multicolumn{1}{|c|}{\textbf{}} & \textbf{eGRU} & \textbf{GRU}  \\ \hline
    \textbf{CoughDetect}            & 95.3\%        & 97.3\%        \\
    \textbf{SpokenDigits}           & 87.8\%        & 91.8\%        \\
    \textbf{UrbanSounds}            & 61.3\%        & 72.1\%        \\ \hline
    Model Size                      & 3 kB          & 34 kB         \\
    Device                          & Arm M0+       & Intel Core i5 \\
    Speed                           & 48 MHz        & 3,800 MHz     \\ \hline
    \end{tabular}

    \vspace{5pt}
    \caption{eGRU deployed on Arm M0+ versus GRU running on Intel workstation.}
    \label{fig:egruperf}
    \end{subfigure}\hspace*{\fill}
    \begin{subfigure}[b]{0.28\textwidth}
    \centering
        \includegraphics[width=\textwidth]{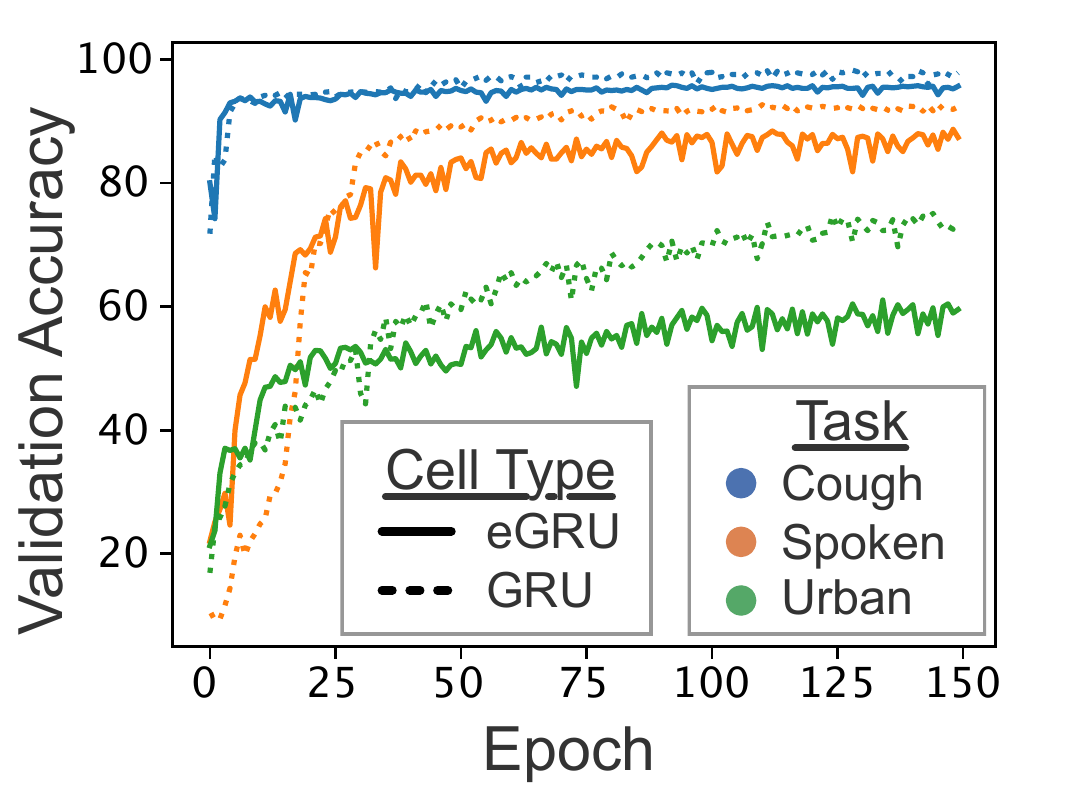}
        \captionof{figure}{Validation accuracy curves for eGRU versus GRU models.}
        \label{fig:valacc}
    \end{subfigure}\hspace*{5pt}
    \begin{subfigure}[b]{0.28\textwidth}
    \centering
        \includegraphics[width=\textwidth]{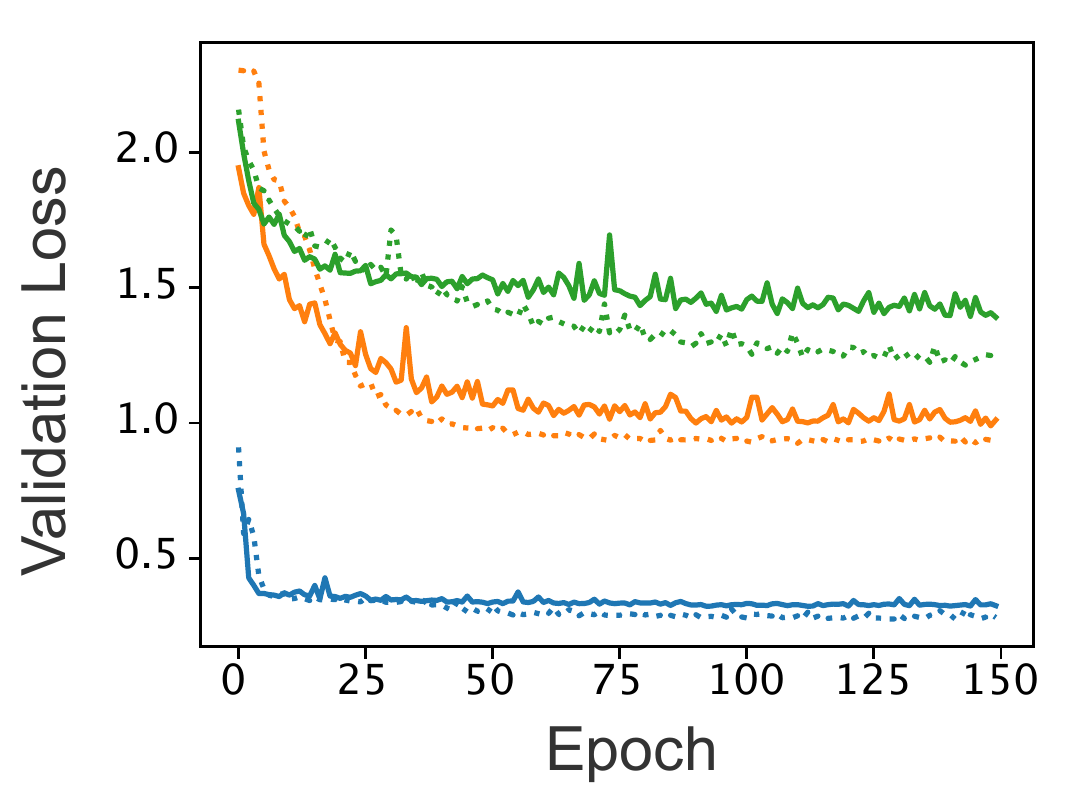}
        \captionof{figure}{Validation loss curves for eGRU versus GRU models.}
        \label{fig:valloss}
    \end{subfigure}    
    \caption{Experiment 4 results showing performance of eGRU model embedded unto the Atmel ATSAMD21 \cite{ATSAMD21}, an ultra low-power Arm M0+ processor. \ref{fig:egruperf} The eGRU model on the M0+ compares well with its full precision GRU counterpart on a computer, across the different task. Yet, the embedded eGRU model is highly efficient, taking only 10\% the size of GRU and requiring inexpensive integer operations exclusively. \ref{fig:valacc} Validation accuracy curves for eGRU and GRU models across all task. eGRU takes longer epochs to train than GRU. \ref{fig:valloss} Corresponding validation loss curves. eGRU models approach similar loss in short-duration tasks.}
    \label{fig:expt4res}
\end{figure}

Finally, in Fig. \ref{fig:expt4res}, the performance of fully optimized eGRU models trained for the different tasks and embedded on the target M0+ device is presented. Note that while eGRU models here feature all optimizations proposed including quantization and fixed point operations, GRU models have full precision weights and employ floating point operations. Also, because a different validation scheme is used, the results for the GRU models here are a bit different than those in the first experiment.

On the embedded system, the eGRU model attains an accuracy of 95.5\% on the cough detection task, which is only 2\% less than the accuracy of a corresponding GRU model on a fully equipped desktop computer. This is impressive considering the eGRU model is only of 3kB in size, 10$\times$ smaller than its GRU equivalent. From latency benchmarks, the embedded eGRU model is measured to take only 9.3 ms to analyze and classify a 350 ms (64 x 24 spectrogram) cough event.

On the spoken digits identification, the eGRU model running on the M0+ still compares well with the GRU. It attains an accuracy of 87.8\% compared to 91.8\% of GRU. That said, the more complex spoken digits task with longer sequences widens the performance difference between eGRU and GRU at 4\%.

Similar to the first experiment, both recurrent cell types perform the worst on the urban sounds dataset. The deployed eGRU model achieves a 61.3\% accuracy, compared to 72.1\% of GRU. Here, the eGRU model under-performs GRU by a significant 11\%. However, that is not very surprising considering the input sequences in this task are 4-10$\times$ longer than in the other two tasks.

\section{Discussion}
The results from the first experiment (Fig. \ref{fig:expt1res} and Fig. \ref{fig:urbansound}) confirm that the proposed architectural modifications in eGRU are indeed effective for short acoustic event detection or keyword spotting tasks. In particular, single gate architecture with softsign activations is seen to be effective for short sequences, and even for longer ones. From a purely architectural stand-point, eGRU is seen to be certainly better than traditional RNN and an efficient candidate for replacing GRU cells. 

With quantization and fixed point arithmetic, eGRU is demonstrated to be efficient enough to fit and run successfully on a resource constrained MCU like the M0+. From the second and third experiments, eGRU is shown to be 10$\times$ smaller and 60$\times$ faster than GRU. Yet, in short sequence tasks, eGRU only suffers 2-4\% reduction in accuracy relative to GRU. Indeed, on more complex AED tasks with much longer sequences such as the urban sounds identification task, fully optimized eGRU under-performs GRU even more. The combination of extreme quantization as well as the single gate architecture appears to heavily impair eGRU's ability to handle long sequences. Thus, eGRU is most suitable for short-duration acoustic event detection applications such as keyword spotting. 

One way to make eGRU effective even for long sequences is to use a less aggressive quantization scheme. As noted in experiment 1 results, full precision eGRU performs well on long sequences even with the single gate architecture. Thus, increasing weight precision from 3 to 4 or even 8 bits will considerably improve accuracy on long events. Another idea for tackling for lengthy sequences is to use eGRU in a sequence-to-sequence network architecture \cite{sutskever2014sequence} with a sequential loss function such as the connectionist temporal classifier (CTC) \cite{graves2006connectionist}. In the sequence-to-sequence with CTC setup, rather than reading an entire input sequence and outputting a single prediction as done above, recurrent networks yield sequential outputs that are aligned with the input. Sequence-to-sequence network architectures are more suitable for long sequences and is typically used in speech recognition tasks. 

However, the significance of eGRU is crucial in ultra-low power wearable applications where resources are limited. For instance, for our wearable cough detector application, the 3 kB eGRU model can easily fit within the 32 kB RAM available on the M0+, with a lot of space left over for other processes. Also, the 9 ms latency can be easily managed on the MCU since cough events are about 350 ms on average. In contrast, the equivalent GRU model requires 34 kB memory and up to 500 ms latency, both of which will not be feasible on the M0+. In such applications, eGRU is the only practical choice.

\section{Conclusion}
In summary, our work explores a means to realize recurrent neural network inferencing on ultra-low power wearable devices. We discussed the constraints of ultra-low-power devices in terms of size, power, memory and computation specifications. We then presented a new recurrent unit architecture, the embedded Gated Recurrent Unit (eGRU), specifically adapted for ultra-low power micro-controller implementation. In particular, eGRU featured a single gate mechanism, faster activation functions, 3-bit weight quantization and utilized exclusively fixed point arithmetic. From our experiments, we demonstrated that while eGRU can be feasibly implemented on target ultra-low power devices, it still performs comparably with GRU on short-duration acoustic event detection applications. 

Two main areas for further investigation remain. First, since final eGRU network only utilized a fraction (10\%) of the available RAM on the M0+, better, bigger and even more complex network architectures can be explored. It will be interesting and beneficial to incorporate recent neural network innovations such as batch normalization, attention mechanisms, and skip/residual connections in the models.

A more pressing next step is to prepare the entire system to run effectively in a real life scenario, untethered and robust to completely unexpected events. Currently, since the datasets used were all created under controlled environments, the model does not perform quite as well on events that are very different from those in the datasets. For the system to be completely untethered, the neural network model has to be combined with other real-time processing steps such as signal acquisition, voice activity detection and spectral decomposition, all running on the ultra-low power micro-controller.

\bibliographystyle{plainnat}

\end{document}